\pdfoutput=1
\documentclass{article}
\usepackage{ijcai24}

\usepackage{times}
\usepackage{soul}
\usepackage{url}
\usepackage[hidelinks]{hyperref}
\usepackage[utf8]{inputenc}
\usepackage[small]{caption}
\usepackage{graphicx}
\usepackage{amsmath}
\usepackage{amsthm}
\usepackage{booktabs}
\usepackage{algorithm}
\usepackage{algorithmic}
\usepackage[switch]{lineno}

\usepackage{newfloat}
\usepackage{listings}
\usepackage{multirow}
\usepackage{amsmath}
\usepackage{amssymb}
\usepackage{bbding}

\title{Boundary-aware Decoupled Flow Networks for Realistic Extreme Rescaling}

\author{
Jinmin Li$^{1,}$\thanks{This work was done while Jinmin Li was an intern at WeChat Pay Lab33, Tencent.}
\and
Tao Dai$^{2,}$\thanks{Corresponding author: Tao Dai (daitao.edu@gmail.com)}
\and
Jingyun Zhang$^4$
\and
Kang Liu$^1$ 
\and
Jun Wang$^4$ 
\and
Shaoming Wang$^4$ 
\and
Shu-Tao Xia$^{1,3}$
\and
Rizen Guo$^4$
\affiliations
$^1$Tsinghua Shenzhen International Graduate School, Tsinghua University
\\
$^2$College of Computer Science and Software Engineering, Shenzhen University\\
$^3$Research Center of Artificial Intelligence, Peng Cheng Laboratory\\
$^4$WeChat Pay Lab33, Tencent\\
\emails
\{ljm22, liuk22\}@mails.tsinghua.edu.cn,
\{daitao.edu, zhang304973926, earljwang\}@gmail.com,
xiast@sz.tsinghua.edu.cn,
\{mangosmwang, rizenguo\}@tencent.com
}

\begin{document}

\maketitle

\begin{abstract}
    Recently developed generative methods, including invertible rescaling network (IRN) based and generative adversarial network (GAN) based methods, have demonstrated exceptional performance in image rescaling. However, IRN-based methods tend to produce over-smoothed results, while GAN-based methods easily generate fake details, which thus hinders their real applications. To address this issue, we propose  Boundary-aware Decoupled Flow Networks (BDFlow) to generate  realistic and visually pleasing results. Unlike previous  methods that model  high-frequency information  as standard Gaussian distribution directly, 
    our BDFlow  first decouples the high-frequency information into \textit{semantic high-frequency} that adheres to a Boundary distribution and \textit{non-semantic high-frequency} counterpart that adheres to a Gaussian distribution. 
    Specifically, to capture semantic high-frequency parts accurately,  we use Boundary-aware Mask (BAM) to constrain the model to produce rich textures, while  non-semantic high-frequency part is randomly sampled from a Gaussian distribution.
    Comprehensive experiments demonstrate that our BDFlow significantly outperforms other state-of-the-art methods while maintaining lower complexity. Notably, our BDFlow improves the PSNR by $4.4$ dB and the SSIM by $0.1$ on average over GRAIN, utilizing only 74\% of the parameters and 20\% of the computation. The code will be available at https://github.com/THU-Kingmin/BAFlow.
\end{abstract}

\begin{figure}[t]
    \centering
    \includegraphics[width=0.8\linewidth]{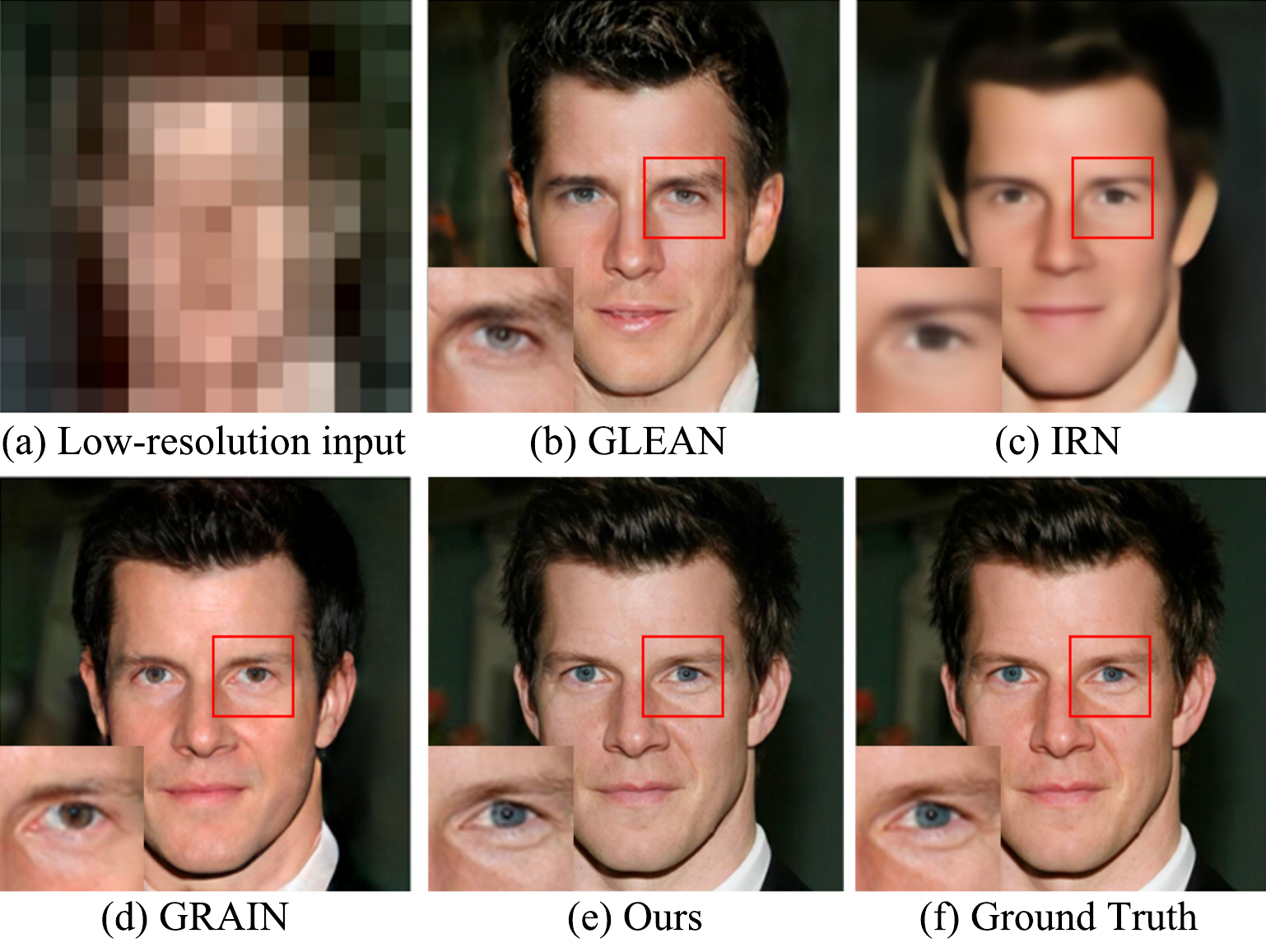}
    \caption{Visual quality of various IRN-based and GAN-based methods, including (b) GLEAN, (c) IRN, and (d) GRAIN. Existing methods produce (c) over-smoothed results or (b) and (d) fake details. By contrast, our method can generate visually pleasing results with sharper details.}
    \label{figs:vis1}
\end{figure}

\section{Introduction}
    Image rescaling,  involving reconstructing high-resolution (HR) images from their corresponding low-resolution (LR) versions, plays a crucial role in large-size data services such as storage and transmission. Early image rescaling methods~\cite{SRCNN,EDSR,RCAN,FSR,CFGN,mambair,cui2023image} primarily concentrated on non-adjustable downscaling kernels and neglected the compatibility between downscaling operations and reconstruction algorithms, which thus easily lose high-frequency information and thus produce visually unpleasing results.
    
    To further improve the reconstruction performance, recently developed invertible rescaling network (IRN) based  and generative adversarial network (GAN) based methods \cite{IRN,GRAIN} have demonstrated impressive performance in image rescaling.
    Among them, IRN~\cite{IRN} transforms high-frequency components into a latent space, and  assumes that the LR components and high-frequency components are independent.
    To consider the effect of the LR counterparts, HCFlow~\cite{HCFlow} 
    using a hierarchical conditional framework for modeling the LR information.
    Later, GRAIN~\cite{GRAIN} introduces a reciprocal invertible image rescaling approach, enabling subtle embedding of HR information into a reversible LR image and generative prior for accurate HR reconstruction.
    Despite the success of these generative methods in image rescaling, they still suffer from visually unpleasing results, due to the inaccurate modeling way of the high-frequency information. As shown in Fig.~\ref{figs:vis1}, existing methods produce over-smoothed results or fake details, thus leading to unrealistic results.
    
\begin{figure}
        \centering
        \includegraphics[width=\linewidth]{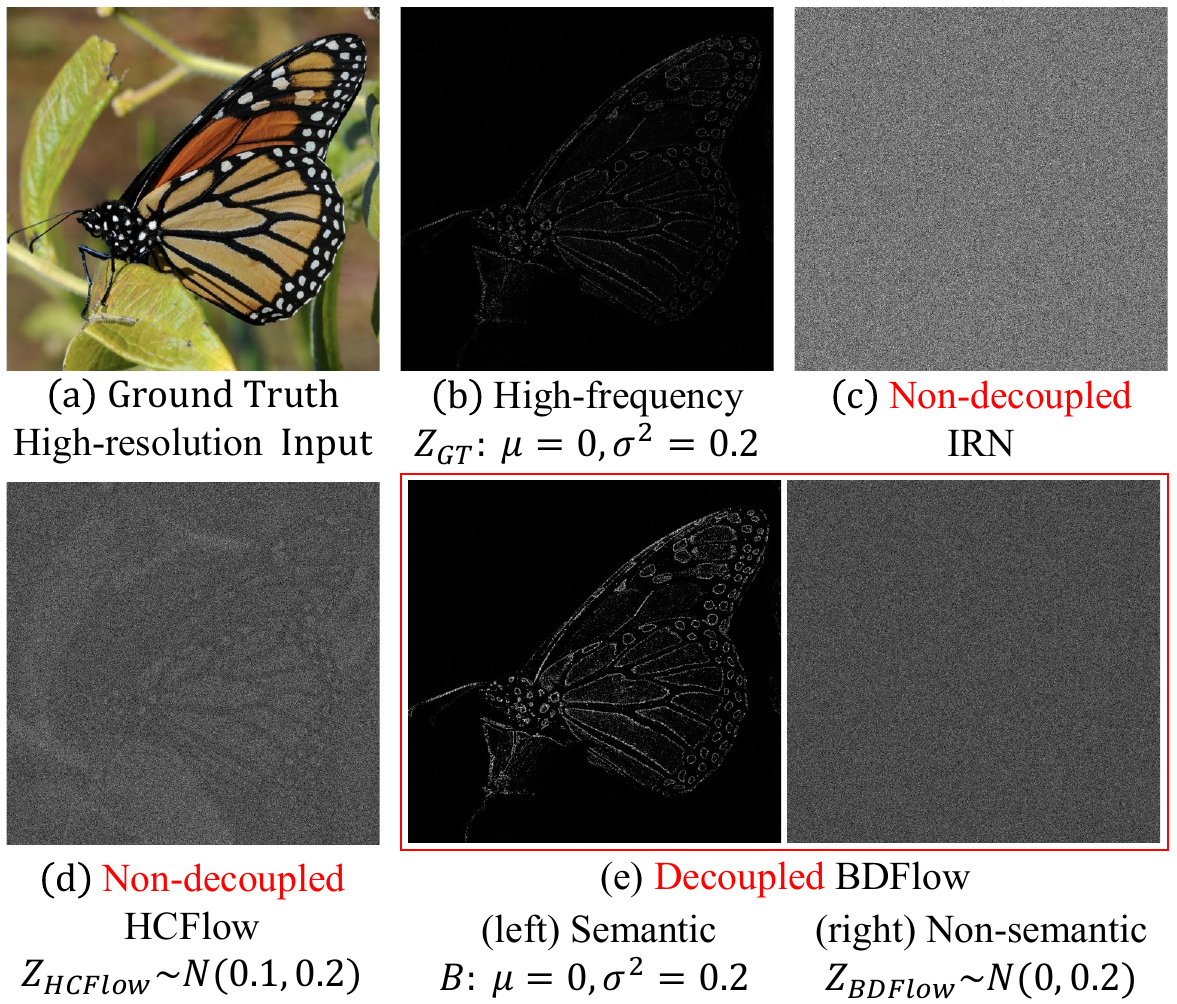}
        \caption{The comparison of different modeling approaches to $Z_{\text{GT}}$. We statistics the distribution of high-frequency information for the DIV2K and CelebA, $Z_{\text{GT}}$: $\mu = 0$, $\sigma^2 = 0.2$. IRN is non-decoupled and models $Z_{\text{GT}}$ as a standard Gaussian distribution, $Z_{\text{IRN}}$. HCFlow is also non-decoupled and learns biased estimation, $Z_{\text{HCFlow}}$. Our BDFlow is decoupled and models the high-frequency information as semantic distribution, $B$ and non-semantic distribution, $Z_{\text{BDFlow}}$.}
        \label{figs:Diff}
    \end{figure}

    On the other hand, it is observed that  the high-frequency part of the input deviates from  a standard Gaussian distribution. 
    As shown in Fig.~\ref{figs:Diff}(b), we compute the statistics of the high-frequency information $Z_{GT}$ from DIV2K and CelebA, which is obtained by computing the difference between the HR image in Fig.~\ref{figs:Diff}(a) and the reconstructed  image of LR. We observe that $Z_{GT}$ is a mixture of semantic boundary information and  Gaussian distribution with $\mu = 0$ and $\sigma^2 = 0.2$.  This motivates us to decouple the high-frequency information into different parts.
    
    The above observations inspire us to design a more effective modeling way for high-frequency information, in this paper,
    we propose Decoupled Boundary-aware Decoupled Flow Networks (BDFlow), which decouples the original high-frequency information into approximate semantic Boundary distribution and non-semantic Gaussian distribution. 
    As shown in Fig.~\ref{figs:vis1}(e), we utilize Boundary-aware Mask (BAM) to preserve semantic information, ensuring that the recovered image follows the true distribution. The non-semantic Gaussian distribution is independent of the low-frequency distribution and semantic Boundary distribution, and is randomly sampled in the backward process. Furthermore, we impose additional constraints on the model to generate textures consistent with the Ground Truth by utilizing the Boundary-aware Weight (BAW). 
    
    In summary, our main contributions are:

    \begin{itemize}
        \item To our best knowledge, the proposed Boundary-aware Decoupled Flow Networks (BDFlow) is the first attempt to decouple high-frequency information into semantic Boundary distribution and non-semantic Gaussian distribution.
        \item We introduce a general Boundary-aware Mask (BAM) to preserve semantic information, ensuring that the recovered image follows the true distribution. 
        Canny operator is a special case of BAM, where the quantization is binarized, and the magnitude of the gradient is calculated using the 2-Norm.
        Additionally, our proposed Boundary-aware Weight (BAW) further constrains the model to generate rich texture details.
        \item Extensive experiments demonstrate that our proposed BDFlow achieves state-of-the-art (SOTA) performance while maintaining a lower computational burden and faster inference time compared to other existing methods. 
        We also explore and analyze image invertibility with respect to different influencing factors, such as scaling factor, JPEG compression, and data domain.
    \end{itemize}
            
\section{Related Work}
    \subsection{Image Rescaling Methods}
        Image rescaling aims to downscale a high-resolution (HR) image to a visually pleasing low-resolution (LR) image and later reconstruct the HR image. In recent years, there has been a surge in the use of normalization flow~\cite{IRN,HCFlow,irrm} and generative adversarial network (GAN)~\cite{GRAIN,GLEAN,GPEN} techniques for tackling the issue of image rescaling. 
        IRN~\cite{IRN} transforms high-frequency information into a latent space $Z\sim p(Z)$ which is a Gaussian distribution. 
        Contrarily, HCFlow\cite{HCFlow} posits that the high-frequency information is dependent on the LR image. 
        GANs-based GRAIN~\cite{GRAIN} introduces a reciprocal approach, enabling delicate embedding of high-resolution information into a reversible low-resolution image and generative prior.
        However, IRN generates overly smooth images due to the substantial loss of high-frequency information. HCFlow is non-decoupled and biased estimation. GRAIN produces fake details and requires extensive computation.
        To address these issues, we introduce light-weight Boundary-aware Mask (BAM) to constrain the model to produce rich textures.
    
    \subsection{Boundary-aware Methods}
        Boundary-aware techniques are vital for preserving facial structure and appearance in image rescaling, especially for facial images. These methods have proven effective in applications like face recognition and segmentation.
        Previous Boundary-aware work involves attention mechanisms~\cite{Attention,HPNet,zhang2023lightweight}, edge-aware filtering~\cite{EdgeAware1,EdgeAware2,li2024fmm}, and facial landmark detection~\cite{Landmark1,Landmark2}. Attention mechanisms focus on important features, improving recognition and segmentation. Edge-aware filtering preserves boundaries while maintaining smoothness elsewhere. Integrating Boundary-aware methods in image rescaling models offers advantages in preserving semantic Boundary distribution and model robustness.

    \subsection{Invertible Residual Networks}
        Initially developed for unsupervised learning of probabilistic models~\cite{InvertibleResidualNetworks,imagegenerate1,imagegenerate2,wang2023contrastive,gao2023backdoor}, Invertible Residual Networks (IRN) facilitate the transformation of one distribution to another through bijective functions, thereby preserving information integrity. This characteristic empowers invertible networks to ascertain the precise density of observations, which can subsequently be employed to generate images with intricate distributions. Owing to these distinctive properties, invertible networks have been effectively utilized in an array of applications, including image rescaling~\cite{IRN,HCFlow}. In this work, we adopt the Invertible Residual Networks architecture presented in~\cite{imagegenerate1}, which consists of a series of invertible blocks. For the $l^{th}$ block, the forward process is formulated as:
        \begin{equation}
            \begin{aligned}
                & u_1^{l+1} = u_1^l + F(u_2^l) \\
                & u_2^{l+1} = u_2^l + G(u_1^{l+1})
            \end{aligned}
            \label{eq:inv}
        \end{equation}

        Here, $F(\cdot)$ and $G(\cdot)$ represent arbitrary transformation functions. The input $u$ is decomposed into $u_1^l$ and $u_2^l$. The backward process can be conveniently defined as:
        \begin{equation}        
            \begin{aligned}
                & u_2^{l} = u_2^{l+1} - G(u_1^{l+1}) \\
                & u_1^{l} = u_1^{l+1} - F(u_2^{l})
            \end{aligned}
        \end{equation}
          
    \begin{figure*}
        \centering
        \includegraphics[width=0.8\linewidth]{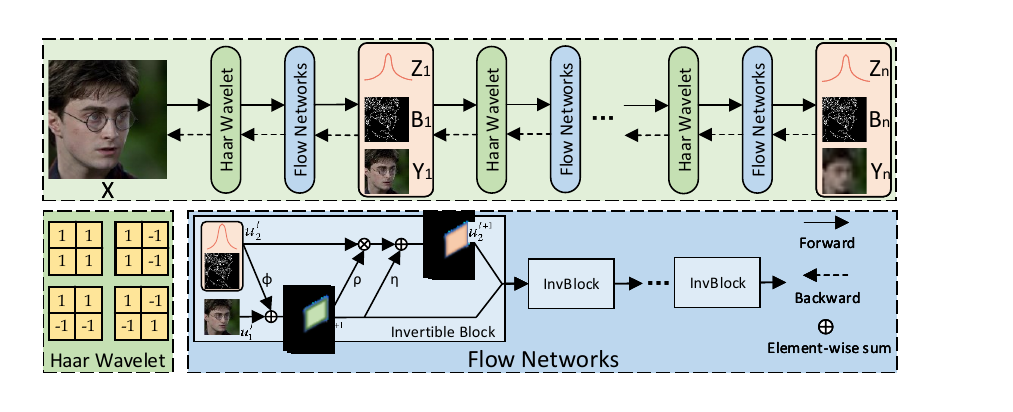}
        \caption{The overall architecture of our  Boundary-aware Decoupled Flow Networks (BDFlow), which comprises Haar Wavelet Blocks and Flow Networks, which further consist of multiple stacked Invertible Blocks (InvBlock). Each InvBlock incorporates three convolutional transformation functions $\phi (\cdot)$, $\rho (\cdot)$, and $\eta (\cdot)$, which enhance the nonlinear representation. $Z_n$ denotes non-semantic high-frequency information that adheres to a Gaussian distribution, while $B_n$ corresponds to semantic high-frequency information that adheres to a Boundary distribution.}
        \label{figs:BDFlow}
    \end{figure*}
\section{Methodology}
    \subsection{Networks Framework}
        BDFlow can be dissected into three components: the Boundary-aware Mask generation algorithm (shown in Algorithm.~\ref{algs:BAM}), the Haar Wavelet transformation~\cite{Haar}, and the Flow Networks (shown in Fig.~\ref{figs:BDFlow}).
        
        Image rescaling focuses on reconstructing a high-resolution (HR) image $X$ from a low-resolution (LR) image $Y$ and high-frequency distribution, which are obtained by downscaling $X$. As the downscaling process is the inverse of upscaling, we employ an invertible neural network to generate the LR image $Y$ and decouple high-frequency distribution to non-semantic Gaussian distribution $Z$ and semantic Boundary distribution $B$ ($i.e., [Y, B, Z] = F_\theta(X), Z \sim p_Z(Z), B \sim p_B(B)$). Inversely, $X$ can be reconstructed through the inverse process from $[Y, B, Z]$: $X = F_\theta^{-1}(Y, B, Z)$. It is important to note that the Boundary distribution preserves semantic information in $B$, and $Z$ corresponds to the non-semantic high-frequency information, as per the Nyquist-Shannon sampling theorem~\cite{NyShan}. To ensure the model's invertibility, we must verify that $F_\theta$ is invertible, which is equivalent to having a non-zero determinant of the Jacobian for each invertible unit $F$.
        \begin{equation}
            J_F =\left[\begin{array}{ll}
            \frac{\partial Y_1}{\partial X_1} & \frac{\partial Y_1}{\partial X_2} \\
            \frac{\partial Y_2}{\partial X_1} & \frac{\partial Y_2}{\partial X_2}
            \end{array}\right]=\left[\begin{array}{cc}
            1 & \frac{\partial F}{\partial X_2} \\
            \frac{\partial G}{\partial X_1} & 1+\frac{\partial G}{\partial F} \frac{\partial F}{\partial X_2}
            \end{array}\right]
        \end{equation}
        
        Here, $F$ and $G$ represent two transformations of each invertible unit. $X$ corresponds to $u^l$, and $Y$ corresponds to $u^{l+1}$ in Eq.~\ref{eq:inv}. The value of $J_F$ is equal to $1$ due to $\frac{\partial G}{\partial X_1}=\frac{\partial G}{\partial F}$.
        
    \subsection{Boundary-aware Decoupled Flow Networks}
        Boundary-aware Decoupled Flow Networks (BDFlow) decouple high-frequency information into non-semantic Gaussian distribution and semantic Boundary distribution. General Boundary-aware Mask preserves semantic information, ensuring that the recovered image follows the true distribution. 
    
        \noindent{\bf Boundary-aware Mask.} The Boundary-aware Mask (BAM) is a crucial component of the BDFlow Networks, and its generation process can be summarized as follows. As shown in Algorithm~\ref{algs:BAM}, given an input image, denoted as $I$, the algorithm begins by applying a Gaussian blur with a standard deviation of $\sigma$ to produce a blurred image $I_g$. Next, the gradient components $G_x$ and $G_y$ of the blurred image $I_g$ are computed, which are then used to calculate the gradient magnitude $M$ (1-Norm, 2-Norm or others). Subsequently, non-maximum suppression is performed on the gradient magnitude $M$ to obtain a thinned boundary map $M'$. The boundary map $M'$ is then sparsified by applying a threshold $T$, resulting in a sparse boundary map $B_s$. Finally, the sparse boundary map $B_s$ is quantified to create the Boundary-aware Mask $B_q$. 

        Canny operator is a special case of BAM, where the quantization is binarized, and the magnitude of the gradient is calculated using the 2-Norm.
        
        \noindent{\bf Boundary-aware Weight.} The proposed Boundary-aware Weight (BAW) further constrains the model to generate rich textures and semantic Boundary distribution, defined as:
        \begin{equation}
            \lambda_{BAW}=\left\{\begin{array}{l}
            \lambda_2 \quad \quad ,\ \text{while}\ E_{cur} < \alpha * E_{max}\\
            \lambda_2 + \frac{B_s - min(B_s)}{max(B_s) - min(B_s)} ,\ \text{otherwise}
            \end{array}\right.
            \label{eq:BAW}
        \end{equation}
        where $\lambda_2$ is a hyper-parameter, $\alpha=0.3$ denotes 30 per cent of the training epochs, $E_{cur}$ is the current training epoch and $E_{max}$ is the max training epoch. During the pre-training period ($\alpha * E_{max}$),  the precision of the BAM is inadequate, and therefore fixed weights are employed. In the later training period, the $B_s$ in the BAM is utilized after normalization, and is used to weight the loss $\mathcal{L}_{back}(X, X_{back})$. The purpose of normalisation is to remove the effects of outliers. $\lambda_{BAW}$ penalises the model for errors in the high-frequency textures of Ground Truth, constraining the model to generate a realistic Boundary distribution.
        
        \begin{algorithm}[tb]
            \caption{Boundary-aware Mask}
            \label{algs:BAM}
            \textbf{Input}: $I$, input image\\
            \textbf{Parameter}: $T$, the threshold to sparsify boundary\\
            \textbf{Output}: $B_q, \text{boundary distribution}$
            \begin{algorithmic}[1] 
                \STATE $I_g \gets \text{GaussBlur}(I, \sigma)$
                \STATE $G_x, G_y \gets \text{Gradient}(I_g)$
                \STATE $M \gets \text{getMagnitude}(G_x, G_y)$
                \STATE $M' \gets \text{NonMaximumSuppress}(M)$
                \STATE $B_s \gets \text{SparsifyBoundary}(M', T)$
                \STATE $B_q \gets \text{Quantify}(B_s)$
                \STATE \textbf{return} $B_q$
            \end{algorithmic}
        \end{algorithm}

        \noindent{\bf Haar Wavelet.} We employ the Haar Wavelet (Other wavelet bases are discussed in the supplementary material.) transformation~\cite{FDWT} to decompose the input $X$ into high and low-frequency information, represented as $[A, H, V, D]$. Specifically, given an HR image $X$ with shape $(H, W, C)$, Haar Wavelet transformation decomposes $X$ into global frequency features $u_1^l, u_2^l$:
        \begin{equation}
            \begin{aligned}
                u_1^l, u_2^l &= \text{HaarWavelet}(X^l)\\
                u_1^l &= [A^l]\\
                u_2^l &= [H^l, V^l, D^l]
            \end{aligned}
        \end{equation}
        where $l$ denotes $l^{th}$ layer of the Flow Networks. $u_1$ and $u_2$ with shape $(\frac{1}{2}H, \frac{1}{2}W, C)$ and $(\frac{1}{2}H, \frac{1}{2}W, 3C)$, correspond to low-frequency and high-frequency information, respectively. 

        \noindent{\bf Flow Networks.} 
        Subsequently, Flow Networks are employed to transform the frequency information into the semantic Boundary distribution, non-semantic Gaussian distribution, and LR distribution. Specifically, wavelet features $u_1^l$ and $u_2^l$ are fed into stacked InvBlocks to obtain the LR $Y$ and high-frequency distribution $(B, Z)$. In our framework, $(B, Z)$ is decoupled into semantic Boundary distribution $B$ and non-semantic Gaussian distribution $Z$. We adopt the general coupling layer for the invertible architecture~\cite{i-Revnet,InvertibleResidualNetworks}. The output of each InvBlock can be defined as:
        \begin{equation}
            \begin{aligned}
                u_1^{l+1} &= u_1^l \cdot  \phi(u_2^l) + \phi(u_2^l) \\
                u_2^{l+1} &= u_2^l \cdot  \rho(u_1^{l+1}) + \eta(u_1^{l+1}) \\
                u_2^l &= (u_2^{l+1} - \eta(u_1^{l+1})) /  \rho(u_1^{l+1}) \\
                u_1^l &= (u_1^{l+1} - \phi(u_2^l)) /  \phi(u_2^l)
            \end{aligned}
        \end{equation}
        where $u_1^{l+1}$ and $u_2^{l+1}$ are the outputs of the current InvBlock and the inputs of the next InvBlock. Notably, $\phi (\cdot)$, $\rho (\cdot)$, and $\eta (\cdot)$ are arbitrary transformation functions. We use the residual block~\cite{ResNet} to enhance the model's nonlinear expressiveness and information transmission capabilities.
        
        InvBlock extracts image features from $u_1^l$ and $u_2^l$ to $u_1^{l+1}$ and $u_2^{l+1}$. After the transformation of $n$ InvBlocks, we obtain the downscaled image $Y$ with dimensions $(\frac{1}{2}H, \frac{1}{2}W, C)$, Boundary distribution $B$ with dimensions $(\frac{1}{2}H, \frac{1}{2}W, 1)$, and Gaussian distribution $Z$ with dimensions $(\frac{1}{2}H, \frac{1}{2}W, 3C-1)$.
        
    \subsection{Loss Function}
    Image rescaling aims to accurately reconstruct the HR image while generating visually pleasing LR images. Following IRN~\cite{IRN} and HCFlow~\cite{HCFlow}, we train our BDFlow by minimizing the following loss:
    \begin{equation}
        \begin{aligned}
            \mathcal{L}= &  \mathcal{L}_{forw}(Y, Y_{forw}) + \mathcal{L}_{back}(X, X_{back})\\
            & + \mathcal{L}_{LPIPS}(X, X_{back})+ \mathcal{L}_{BAM}(B, B_{forw}) \\
            & + \mathcal{L}_{latent}(Z)
        \end{aligned}
    \end{equation}
    where $X$ and $Y$ are the ground-truth HR image and LR image, respectively. $B$ and $Z$ are the ground-truth semantic Boundary distribution and non-semantic Gaussian distribution, respectively.$X_{back}$ is the reconstructed HR image from the generated LR image $Y_{forw}$, Boundary distribution $B_{forw}$, and sampled Gaussian distribution $Z$.
    
    $\mathcal{L}_{forw}$ is the $l_2$ pixel loss defined as:
    \begin{equation}
        \mathcal{L}_{forw}(Y, Y_{forw}) = \lambda_1 \frac{1}{N} \sum_{i=1}^N\|F_\theta^Y(X)-Y\|_2
    \end{equation}
    where $N$ is the number of pixels, $Y_{forw} = F_\theta^{Y}(X)$ is the generated LR image, and $\lambda_1$ is a hyper-parameter balancing different losses.

    $\mathcal{L}_{back}$ is the $l_1$ pixel loss defined as:
    \begin{equation}
        \mathcal{L}_{back}(X, X_{back}) = \lambda_{BAW} \frac{1}{M} \sum_{i=1}^M\|F_\theta^{-1}(Y, B, Z)-X\|_1
    \end{equation}
    where $M$ is the number of pixels, $X_{back} = F_\theta^{-1}(Y, B, Z)$, and $\lambda_{BAW}$ is generated by the generation process of BAM but does not need to be quantified, which is $B_s$ in Algorithm~\ref{algs:BAM} and Eq.~\ref{eq:BAW}.
    
    $\mathcal{L}_{LPIPS}$ with the weight $\lambda_3$ is used to enhance the visual effect of the image generated by the model. 
    Similarly, $\mathcal{L}_{BAM}$ is the $l_2$ pixel loss defined as:
    \begin{equation}
        \mathcal{L}_{BAM}(B_{forw}, B)= \lambda_4 \frac{1}{N} \sum_{i=1}^N\|F_\theta^B(X)-B\|_2
    \end{equation}
    where $N$ is the number of pixels, $\lambda_4$ is a hyper-parameter, and $B_{forw} = F_\theta^{B}(X)$ is the generated semantic Boundary distribution.
    
    The last term, $\mathcal{L}_{latent}(Z)$, chooses more stable distribution metrics for minimization to ensure that $Z$ follows a non-semantic Gaussian distribution~\cite{IRN}, defined as:
    \begin{equation}
        \mathcal{L}_{latent}(Z) = - \lambda_5 \mathbb{E}_{q(X)}\left[\log p\left(Z=F_\theta^Z(X)\right)\right]
    \end{equation}
    
    We jointly optimize the invertible architecture $F$ by utilizing both forward and backward losses.

\section{Experiments}
    \subsection{Datasets}
        We utilize the CelebA-HQ dataset~\cite{CelebAHQ} to train our BDFlow. The dataset comprises 30,000 high-resolution ($1024 \times 1024$) human face images. Moreover, we evaluate our models using widely-accepted pixel-wise metrics, including PSNR, SSIM, and LPIPS~\cite{LPIPS} (Y channel) on the CelebA-HQ test dataset. We additionally train our BDFlow on the Cat dataset~\cite{CAT} and the LSUN-Church~\cite{LSUN} dataset to evaluate its generalization ability across different domains. 
    
    \subsection{Training Details}
        Our BDFlow consists of $n$ Haar Wavelet Transformation and Flow Networks for $2^n\times$, each containing two Invertible Blocks (InvBlocks). We train these models using the ADAM \cite{ADAM} optimizer with $\beta_1 = 0.9$ and $\beta_2 = 0.999$. Furthermore, we initialize the learning rate at $2 \times 10^{-4}$ and apply a cosine annealing schedule, decaying from the initial value to $1 \times 10^{-6}$ over the total number of iterations. We set $\lambda_1$, $\lambda_2$, $\lambda_3$, $\lambda_4$, and $\lambda_5$ to 2, 2, 1, 16, and 4, respectively.
    
    \begin{table}[t]
        \small
        \centering
        \begin{tabular}{c|c|ccc}
            \hline \multirow{2}{*}{ Scale } & \multirow{2}{*}{ Model } & \multicolumn{3}{|c}{ CelebA } \\
            \cline{3-5} & & PSNR & SSIM  & LPIPS \\
            \hline \multirow{5}{*}{$\times 8$} & Bicubic \& DIC & 25.55 & 0.7574 & 0.5526 \\
             & Bicubic \& SRFlow & 26.74 & 0.7600 & 0.2160 \\
             & HCFlow & 26.66 & 0.7700 & 0.2100 \\
             & BDFlow (ours) & $\bf {38.93}$ & $\bf{0.9425}$ & $\bf{0.0850}$ \\
            \hline \multirow{4}{*}{$\times 16$} & Bicubic \& WSRNet & 22.91 & 0.6201 & 0.5432 \\
             & Bicubic \& ESRGAN & 21.01 & 0.5959 & 0.4464 \\
             & BDFlow (ours) & $\bf{31.97}$ & $\bf{0.8219}$ & $\bf{0.2476}$ \\
            \hline \multirow{4}{*}{$\times 32$} & GPEN & 20.40 & 0.5919 & 0.3714 \\
             & IRN & 24.41 & 0.6943 & 0.5238 \\
             & BDFlow (ours) & $\bf{29.15}$ & $\bf{0.8060}$ & $\bf{0.3273}$ \\
            \hline \multirow{7}{*}{$\times 64$} & PULSE & 19.20 & 0.5515 & 0.4867 \\
             & pSp & 17.70 & 0.5590 & 0.4456 \\
             & Bicubic \& GLEAN & 20.24 & 0.6354 & 0.3891 \\
             & Bicubic \& Bilinear & 19.92 & 0.6840 & 0.6027 \\
             & TAR & 25.15 & 0.7397 & 0.4733 \\
             & GRAIN & 22.30 & 0.6467 & $\bf{0.2686}$ \\
             & BDFlow (ours) & $\bf{26.12}$ & $\bf{0.7805}$ & 0.3672 \\
            \hline
        \end{tabular}
        \caption{Quantitative comparison results (PSNR / SSIM / LPIPS) of various face-rescaling methods on CelebA are presented. The best results are highlighted in bold.}
        \label{tabs:SOTA}
    \end{table}

    \begin{figure*}
        \centering
        \includegraphics[width=0.8\linewidth]{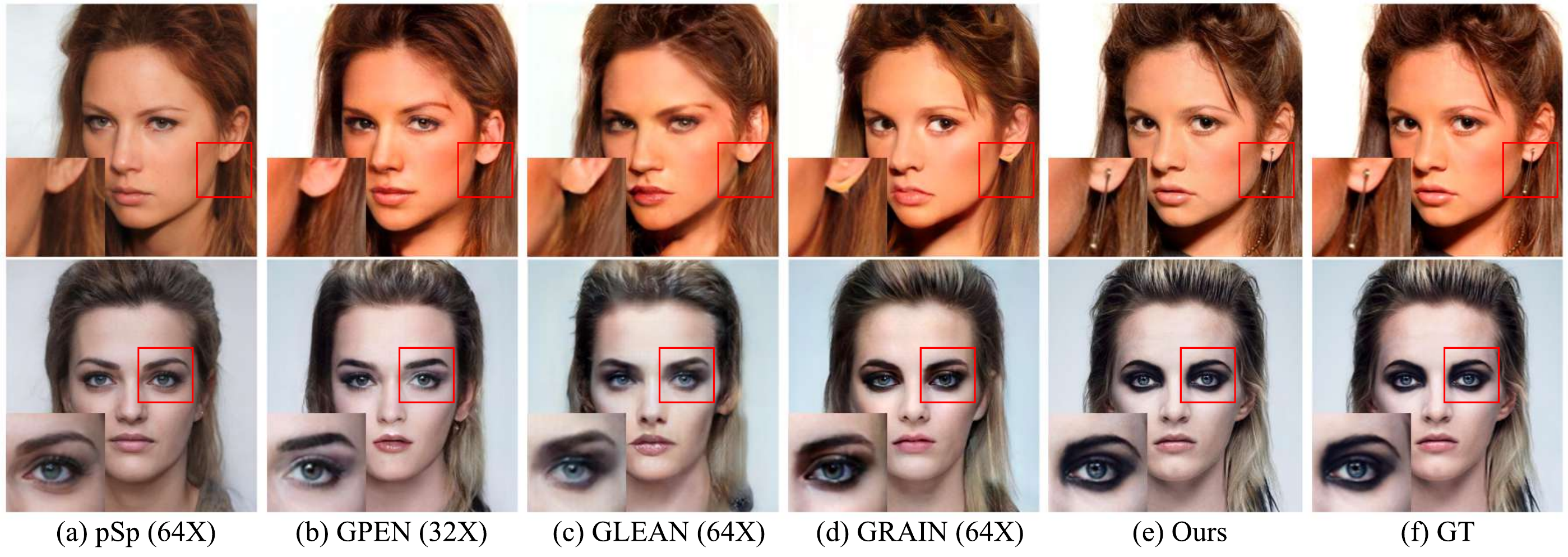}
        \caption{Visual results of rescaling the HR images with $1024 \times 1024$. The LR images are $16 \times 16$ for rescaling factor $64$. BDFlow recovers rich textures and realistic details, leading to better recovery performance.}
        \label{figs:vis2}
    \end{figure*}
    
    \subsection{Comparison with State-of-the-art Methods}
        We compare our BDFlow with several state-of-the-art methods, including GAN-based inversion methods, CNN-based SR methods, and invertible rescaling methods. Table~\ref{tabs:SOTA} presents the quantitative evaluations of the various approaches. Evidently, CNN-based methods outperform GAN-based methods in terms of PSNR and SSIM, as they tend to produce fuzzy results that follow the overall structure. Our flow-based approach BDFlow achieves superior PSNR and SSIM, with LPIPS results second only to GRAIN~\cite{GRAIN} on $64\times$. This demonstrates that our model considers not only pixel-level fidelity but also perceptual visual effects in facilitating the decoupling of high-frequency information into semantic Boundary distribution and non-semantic Gaussian distribution. Specifically, BDFlow improves the PSNR by $4.4~dB$ and the SSIM by $0.0976$ on average over GRAIN.

        Table~\ref{tab:FLOPs_Time} compares IRN~\cite{IRN}, GRAIN~\cite{GRAIN}, and our BDFlow in terms of model parameters, computational effort, and inference time. Our BDFlow surpasses IRN in all dimensions and outperforms GRAIN by using only 74\% of the parameters and 20\% of the computation. Notably, BDFlow is suitable for real-time applications, as it can achieve 100 fps.

        Fig.~\ref{figs:vis2} contrasts our BDFlow with the GAN-based approach. In particular, pSp~\cite{pSp} recovers images of faces with different identities, while the results of GPEN~\cite{GPEN} and GLEAN~\cite{GLEAN} display slight quality improvements but still exhibit significant flaws. Major differences persist in hair, eyebrows, eyes, and facial expressions. Although GRAIN~\cite{GRAIN} combines reversible and generative priors, it can only mitigate the limitation of under-informed input. GRAIN achieves good detail generation but produces some unfaithful details, as shown in Fig.~\ref{figs:vis2} (d). In contrast, as shown in Fig.~\ref{figs:vis2} (e), our BDFlow excels in both realism and fidelity, capturing semantic facial features, like vivid facial expressions, flyaway hair, and eye color.

        \begin{table}
            \small
            \centering
            \begin{tabular}{c|c|c|c|c}
                \hline Scale & Model & Paras $(\mathrm{M})$  & FLOPs $(\mathrm{G})$  & Time $(\mathrm{s})$  \\
                \hline \multirow{2}{*}{$\times 8$} & IRN & $11.12 $ & $11.2965 $ & $0.146 $ \\
                 & BDFlow & $\bf{3.16}$ & $\bf{3.4399}$ & $\bf{0.050}$ \\
                \hline \multirow{2}{*}{$\times 16$} & IRN & $34.19 $ & $12.7721 $ & $0.149 $ \\
                 & BDFlow & $\bf{4.58}$ & $\bf{1.9119}$ & $\bf{0.027}$ \\
                \hline \multirow{2}{*}{$\times 32$} & IRN & $122.49 $ & $14.1814 $ & $0.186 $ \\
                 & BDFlow & $\bf{15.94}$ & $\bf{2.0933}$ & $\bf{0.032}$ \\
                \hline \multirow{3}{*}{$\times 64$} & IRN & $471.72 $ & $15.5802 $ & $0.227 $ \\
                 & GRAIN & $82.27 $ & $11.5963 $ & $3.645 $ \\
                 & BDFlow & $\bf{60.72}$ & $\bf{2.2720}$ & $\bf{0.036}$ \\
                \hline
            \end{tabular}
            \caption{Quantitative comparison results (Parameters, FLOPs, Time) among IRN, GRAIN and our BDFlow on CelebA. The best results are indicated in bold.}
            \label{tab:FLOPs_Time}
        \end{table}
    
    \subsection{Ablation Study}
        We conducted an ablation study to evaluate the impact of our proposed BAM and BAW. The effects of LPIPS loss, different Norm Loss, different wavelet bases, and hyper-parameters are discussed in the supplementary material.
        \begin{table}[]
            \small
            \centering
            \begin{tabular}{c|c|c|c|c}
                \hline BAM & BAW & PSNR $\uparrow$ & SSIM $\uparrow$ & LPIPS $\downarrow$ \\
                \hline \CheckmarkBold & \XSolidBrush & 26.06 & 0.7798 & 0.3673 \\
                \hline \XSolidBrush &\CheckmarkBold & 25.03 & 0.7652 & 0.3823 \\
                \hline \CheckmarkBold & \CheckmarkBold & $\mathbf{26.12}$ & $\mathbf{0.7805}$ & $\mathbf{0.3672}$ \\
            \hline
            \end{tabular}
            \caption{Ablation study evaluating the impact of the Boundary-aware Mask and Boundary-aware Weight in the CelebA with $\times 64$.}
            \label{tab:BAMBAW}
        \end{table}

        \begin{table}[]
            \small
            \centering
            \begin{tabular}{c|c|c|c|c|c}
                \hline Quantify & Magnitude & T & PSNR & SSIM  & LPIPS \\
                \hline 1 bit & 2-Norm & 20 & 26.09 & 0.7799 & 0.3681 \\
                 1 bit & 2-Norm & 50 & $\mathbf{2 6 . 1 2}$ & $\mathbf{0 . 7 8 0 5}$ & $\mathbf{0 . 3 6 7 2}$ \\
                 1 bit & 2-Norm & 100 & 26.11 & 0.7802 & 0.3675 \\
                \hline 1 bit & 1-Norm & 50 & 26.09 & 0.7798 & 0.3684 \\
                 1 bit & 2-Norm & 50 & 26.12 & 0.7805 & 0.3672 \\
                 2 bit & 2-Norm & 50 & 26.51 & 0.7913 & 0.3495 \\
                 3 bit & 2-Norm & 50 & $\mathbf{2 7 . 1 3}$ & $\mathbf{0 . 8 2 1 1}$ & $\mathbf{0 . 3 2 6 3}$ \\
                \hline
            \end{tabular}
            \caption{Ablation study evaluating the impact of quantify level and calculation of magnitude in the CelebA with $\times 64$. T denotes the magnitude threshold.}
            \label{tab:BAM}
        \end{table}

\begin{figure}[t]
            \centering
            \includegraphics[width=\linewidth]{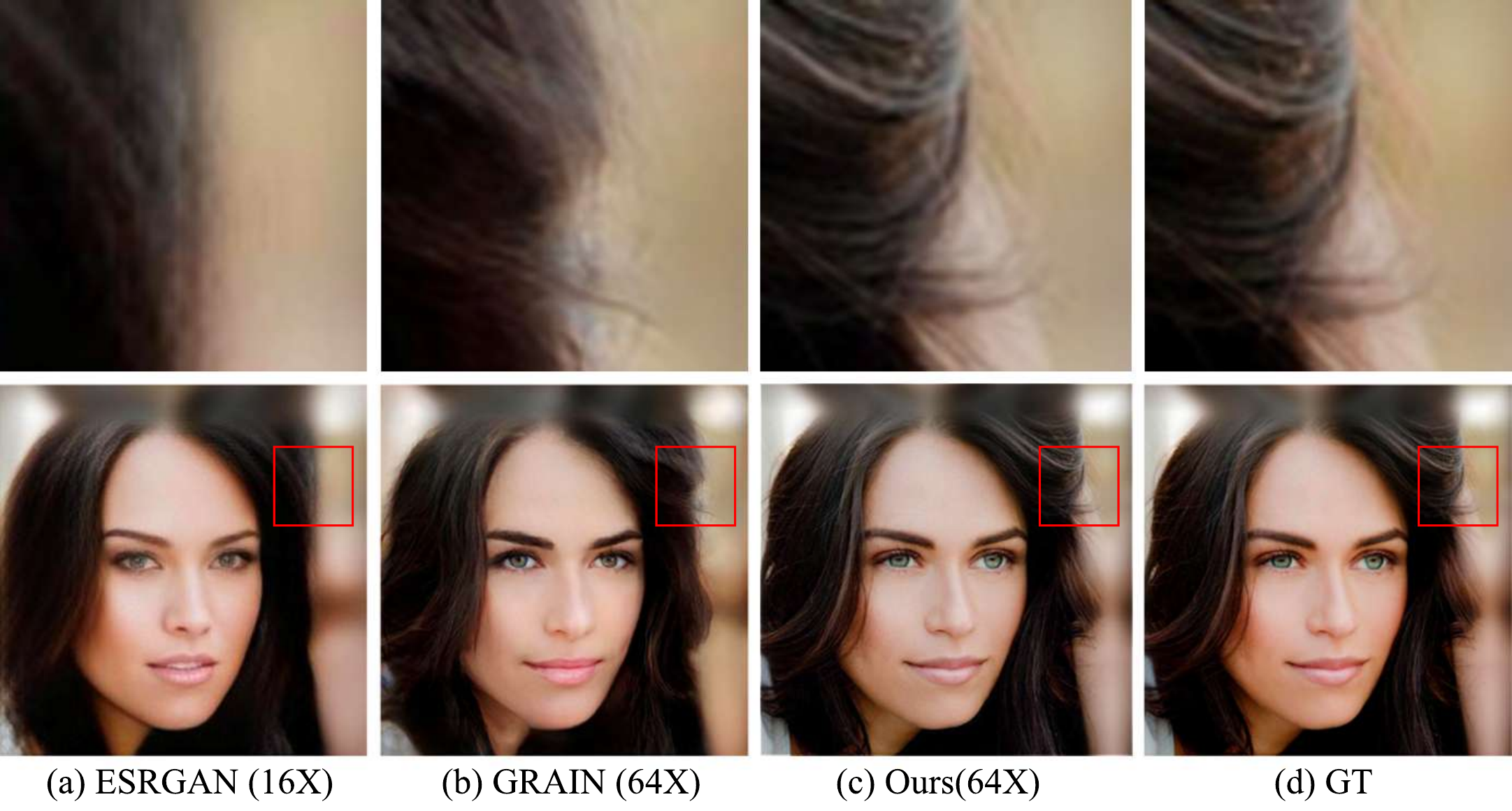}
            \caption{Visual results of rescaling the HR images with $1024 \times 1024$. The LR images are $16 \times 16$ for $\times 64$. BDFlow helps to restore more realistic and clear hair details.}
            \label{figs:vis3}
        \end{figure}

        \noindent{\bf Boundary Distribution and Semantic High-frequency Textures.} Boundary-aware Mask (BAM) and Boundary-aware Weight (BAW) are able to recover more realistic Boundary distribution, such as the hair strands shown in Fig.~\ref{figs:vis3} (c), because they focus on handling image boundaries and regions with high-frequency information more effectively. This is achieved by encouraging the model to decouple the high-frequency information into non-semantic Gaussian distribution and semantic Boundary distribution. However, GAN-based methods, like ESRGAN~\cite{ESRGAN} and GRAIN~\cite{GRAIN}, generate fake details without the constraint of semantic Boundary distribution. 
        
        Table~\ref{tab:BAMBAW} demonstrates that the BDFlow structure incorporating both BAM and BAW consistently outperforms the basic BDFlow structure in terms of PSNR, SSIM, and LPIPS. Specifically, Table~\ref{tab:BAM} shows that BAM with a threshold value of $T=50$ achieves superior performance due to the smaller values of high-frequency textures, which have a mean of approximately 50. Additionally, using the 2-Norm for magnitude calculation is better than using the 1-Norm because it widens the gap between different magnitudes. It is worth noting that the Canny operator can be regarded as a special case of BAM, where quantization is binarized and the magnitude of the gradient is calculated using the 2-Norm. However, as the level of quantization increases, the model achieves better performance. The Boundary distribution with 1 bit and single channel requires only a 4\% ($\frac{1}{8} \times \frac{1}{3} \approx 4\%$) increase in storage cost, but can be further compressed to 1\% due to the sparsity of high frequency.
           
    \subsection{Discussions}
        In our discussions, we conducted various analyses to evaluate the effectiveness of our approach. Firstly, we conducted a comparative analysis against the widely-used JPEG image compression technique to establish its superiority. Next, we evaluated the effect of BDFlow on different domains. More discussion is placed in the supplementary material, including the scale of sampled $Z$ and evaluation of LR Images.
            \begin{table}
                \centering
                \begin{tabular}{c|c|cc}
                    \hline \multirow{2}{*}{ Quality } & \multirow{2}{*}{ Model } & \multicolumn{2}{|c}{ CelebA } \\
                    \cline { 3 - 4 } & & PSNR (dB) $\uparrow$ & Storage (B) $\downarrow$ \\
                    \hline 1 & JPEG & 23.64 & 18803 \\
                     $16 \times 16$ & GRAIN & 22.30 & $\bf{724}$ \\
                     $16 \times 16$ & BDFlow & $\bf{26.12}$ & $\bf{724}$ \\
                    \hline 5 & JPEG & 26.94 & 21511 \\
                     $32 \times 32$ & GRAIN & 25.60 & $\bf{978}$ \\
                     $32 \times 32$ & BDFlow & $\bf{29.15}$ & $\bf{978}$ \\
                    \hline 10 & JPEG & 31.21 & 27933 \\
                     $64 \times 64$ & GRAIN & 28.13 & $\bf{1731}$ \\
                     $64 \times 64$ & BDFlow & $\bf{31.97}$ & $\bf{1731}$ \\
                    \hline 40 & JPEG &  $\bf{38.99}$ & 52979 \\
                     $128 \times 128$ & GRAIN & 32.60 & $\bf{4103}$ \\
                     $128 \times 128$ & BDFlow & 38.93 & $\bf{4103}$ \\
                    \hline
                \end{tabular}
                \caption{Quantitative comparison with JPEG compression technology and GRAIN in PSNR and storage size. Quality indicates the level of JPEG compression, and $n \times n$ denotes the input size of the LR image.}
                \label{tab:JPEG}
            \end{table}

        \noindent{\bf Comparison with JPEG Compression.} Table~\ref{tab:JPEG} illustrates the superior performance of BDFlow compared to JPEG compression and GRAIN~\cite{GRAIN} in terms of PSNR and storage size. BDFlow consistently achieves higher PSNR values while maintaining storage sizes comparable to GRAIN, showcasing its ability to effectively compress images while preserving the quality of the reconstructed images. BDFlow's decoupling of high-frequency information into semantic Boundary distribution and non-semantic Gaussian distribution is at the core of the above gains. 
        In addition, JPEG compression and rescaling are not in conflict, it is possible to continue to compress images using the JPEG algorithm after rescaling.
        \begin{table}[h]
                \centering
                \begin{tabular}{c|c|c|c}
                    \hline Domain & Model & PSNR $\uparrow$ & SSIM $\uparrow$ \\
                    \hline \multirow{2}{*}{ Cat } & GRAIN ($\times 32$) & 21.97 & 0.5250 \\
                     & BDFlow ($\times 32$) & $\mathbf{27.47}$ & $\mathbf{0.7538}$ \\
                    \hline \multirow{2}{*}{ Church } & GRAIN ($\times 32$) & 18.48 & 0.4421 \\
                     & BDFlow ($\times 32$) & $\mathbf{25.31}$ & $\mathbf{0.7489}$ \\
                    \hline \multirow{2}{*}{ Set5 } & GRAIN ($\times 64$) & 22.33 & 0.7718 \\
                     & BDFlow ($\times 64$) & $\mathbf{24.80}$ & $\mathbf{0.8361}$ \\
                    \hline \multirow{2}{*}{ Set14 } & GRAIN ($\times 64$) & 19.96 & 0.6055 \\
                     & BDFlow ($\times 64$) & $\mathbf{22.07}$ & $\mathbf{0.6784}$ \\
                    \hline \multirow{3}{*}{ B100 } & GRAIN ($\times 64$) & 21.56 & 0.6128 \\
                     & BDFlow ($\times 64$) & $\mathbf{23.37}$ & $\mathbf{0.6923}$ \\
                    \hline \multirow{2}{*}{ Urban100 } & GRAIN ($\times 64$) & 16.77 & 0.3886 \\
                     & BDFlow ($\times 64$) & $\mathbf{18.02}$ & $\mathbf{0.4560}$ \\
                    \hline \multirow{2}{*}{ DIV2K } & GRAIN ($\times 64$) & 19.03 & 0.4901 \\
                     & BDFlow ($\times 64$) & $\mathbf{19.89}$ & $\mathbf{0.5114}$ \\
                    \hline
                \end{tabular}
                \caption{Quantitative comparisons of different domains in Cat, Church, Set5, Set14, B100, Urban100 and DIV2K.}
                \label{tab:Domain}
            \end{table}
         \begin{figure}[t]
                \centering
                \includegraphics[width=\linewidth]{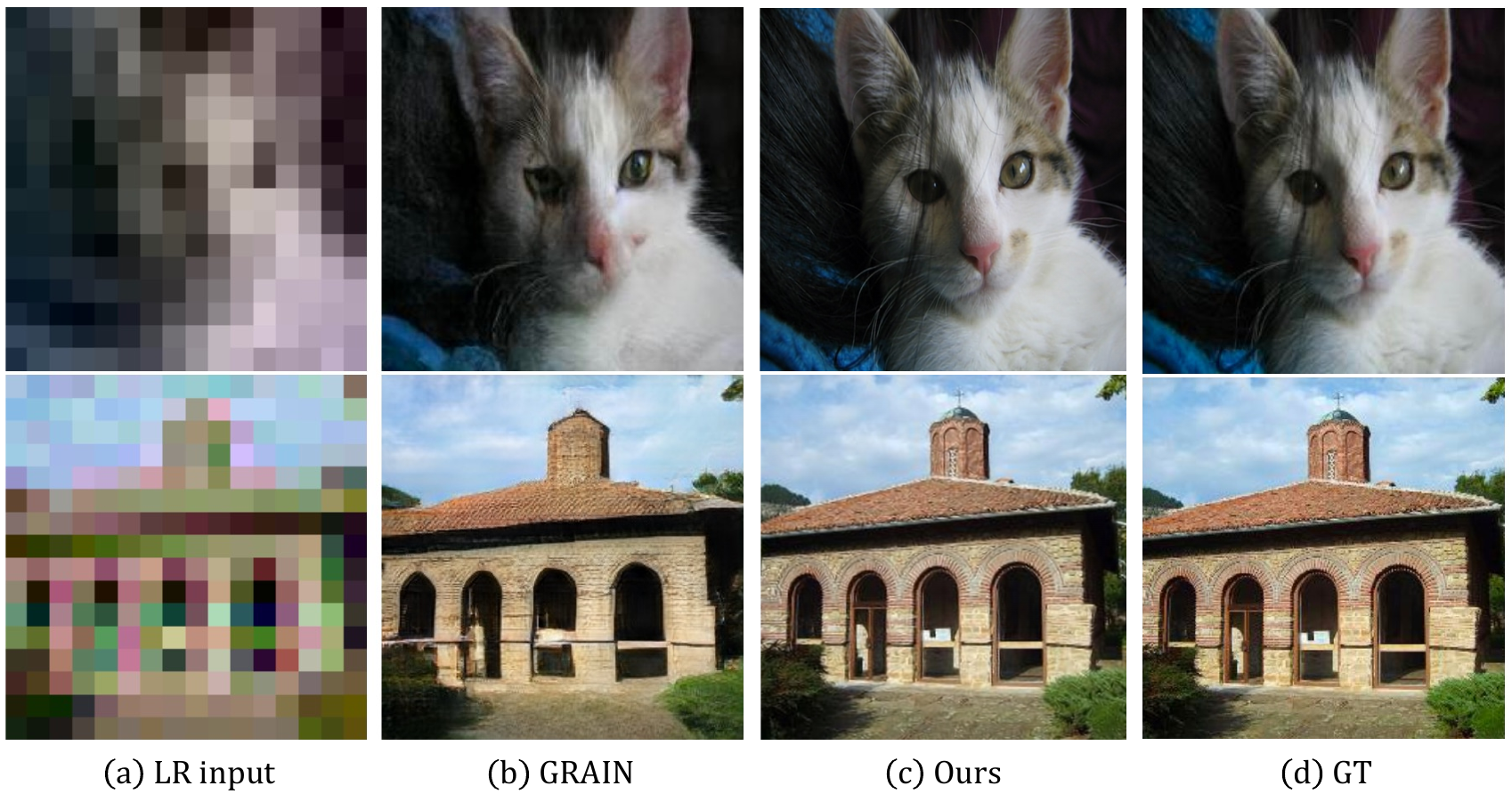}
                \caption{Visual results of our BDFlow on Cat domain and church domain with $\times 16$. BDFlow helps to reconstruct more texture details and obeys the true distribution.}
                \label{figs:Domain}
            \end{figure}
            
        \noindent{\bf Evaluation in Different Domains.} As shown in Fig.~\ref{figs:Domain} and Table~\ref{tab:Domain}, we evaluate our proposed approach in different Domains. These experiments demonstrate the generalisation ability of our proposed BDFlow in reconstructing images from different Domains. This is attributed to the fact that we decouple the true high-frequency distributions into semantic Boundary distribution and non-semantic Gaussian distribution, which helps to achieve the final realistic results. As shown in Fig.~\ref{figs:Domain}, GRAIN has difficulty recovering the hair of the cat and the edges of the church, but our BDFlow is able to do so with high fidelity. As shown in Table~\ref{tab:Domain}, our BDFlow achieves much higher PSNR and SSIM than GRAIN at the pixel level.

        \noindent{\bf $l_1$ Loss and $l_2$ Loss.} We use $l_1$ loss and $l_2$ loss in some cases in our method for the fairness comparison, since we followed the default setup of the previous works \cite{IRN,HCFlow}. 
        Besides, we have conducted experiments to verify the effects of  Norm-loss (see in Table 1 of the Supplementary Material). In fact,  the forward process utilizes $l_2$ to increase the penalty for outliers in favor of constraining the LR distribution, Gaussian distribution, and Boundary distribution. The backward process utilizes the $l_1$ loss encouraging the model to recover more detail rather than smoothing the image, thus leading to better LPIPS. Thus, $(l_2, l_1)$ may be a proper choice.   
        
\section{Conclusion}
    In this paper, we proposed the Boundary-aware Decoupled Flow Networks (BDFlow), which decouple high-frequency information into non-semantic Gaussian distribution and semantic Boundary distribution. 
    This approach effectively preserves boundary information and texture details in the image rescaling process. Specifically, by introducing a generated Boundary-aware Mask, our model ensures that the recovered image follows the true high-frequency distribution and generates rich semantic details. Boundary-aware Mask further constrains the generation of the model to obey the true semantic distribution. 

\section{Acknowledgements}
    This work is supported in part by the National Natural Science Foundation of China, under Grant (62302309, 62171248),  Shenzhen Science and Technology Program (JCYJ20220818101014030, JCYJ20220818101012025), and the PCNL KEY project (PCL2023AS6-1), and Tencent ``Rhinoceros Birds" - Scientific Research Foundation for Young Teachers of Shenzhen University.
    
\bibliographystyle{named}
\bibliography{main}

\end{document}